# Accounting for Context in Plan Recognition, with Application to Traffic Monitoring

David V. Pynadath and Michael P. Wellman
Artificial Intelligence Laboratory
University of Michigan
1101 Beal Avenue
Ann Arbor, MI 48109-2110 USA
{pynadath, wellman}@engin.umich.edu

## Abstract

Typical approaches to plan recognition start from a representation of an agent's possible plans, and reason evidentially from observations of the agent's actions to assess the plausibility of the various candidates. A more expansive view of the task (consistent with some prior work) accounts for the context in which the plan was generated, the mental state and planning process of the agent, and consequences of the agent's actions in the world. We present a general Bayesian framework encompassing this view, and focus on how context can be exploited in plan recognition. We demonstrate the approach on a problem in traffic monitoring, where the objective is to induce the plan of the driver from observation of vehicle movements. Starting from a model of how the driver generates plans, we show how the highway context can appropriately influence the recognizer's interpretation of observed driver behavior.

## 1 INTRODUCTION

The problem of *plan recognition* is to induce the plan of action driving an agent's behavior, based on partial observation of its behavior up to the current time. Deriving the underlying plan can be useful for many purposes—predicting the agent's future behavior, interpreting its past behavior, or generating actions designed to influence the plan itself. Researchers in AI have studied plan recognition for several kinds of tasks, including discourse analysis (Grosz & Sidner, 1990), collaborative planning (Huber & Durfee, 1993), and adversarial planning (Azarewicz *et al.*, 1989). These works have employed a great variety of reasoning techniques, operating on similarly various plan representations and adopting varied assumptions about observability.

The common theme underlying these diverse motivations and approaches is that the object to be induced is a *plan*, and that this plan is the cause of observed behavior. If there is anything special about the task of plan recognition as opposed to recognition in general, it must be due to special properties of plans: how they are constituted, and how they cause the behavior we observe and wish to predict, interpret, and influence.

In this paper, we focus on one of these special properties—the context in which the plan is generated—and how it can be exploited in the recognition process. Whereas most previous approaches have emphasized the relationship between plans and their observable effects,[1] we argue that it is equally necessary to consider evidence that would bear on which plan would have been appropriate for the agent to generate. We demonstrate this point through an example application in traffic monitoring, where the interpretation of an individual vehicle's action depends on the surrounding highway context. Our techniques for reasoning about plan-generation context are based on Bayesian networks, as part of a general Bayesian framework for plan recognition. This contribution can be considered a variant extension of the model of Charniak and Goldman (1993), and of the approach advocated by Huber et al. (1994).

## 2 PLAN RECOGNITION

### 2.1 TOWARDS A GENERAL BAYESIAN FRAMEWORK

One of the aims of our work is to elucidate the fundamental elements of plan recognition, and to develop a general Bayesian framework for approaches to this task. Achieving generality is complicated by the diversity of representations for plans and techniques for plan generation; therefore, we present the framework at multiple levels of specificity. The most abstract specification is designed to accommodate most conceivable versions of plan recognition, and by introducing further distinctions we taxonomize the ap-proaches.

---

[1] Although, as we point out in the discussion below, several of these approaches can also accommodate the sort of context information we are concerned with.

The framework for plan recognition is distinguished from uncertain reasoning in general by two special features of plans. First, plans are *structured linguistic objects*. Plan languages considered in AI research range from simple sequences of action tokens to general-purpose programming languages. In either case, the recognizer can and should exploit the structure of plans in inducing them from partial observations of the actions comprising the plan. Another way to say this is that plans are descriptions of action *patterns*, and therefore any general pattern-recognition technique is automatically a plan recognition technique for the class of plans corresponding to the class of patterns associated with the given technique.

The second special feature of plans is that they are *rational constructions*. They are synthesized by a rational agent with some beliefs, preferences, and capabilities, that is, a *mental state*. Knowing the agent's mental state and its rationality properties strongly constrains the possible plans it will construct. (The degree of constraint depends on the power of the rationality theory we adopt.) The rational origin of plans is what distinguishes plan recognition from pattern recognition. If the observations available include evidence bearing on the beliefs, preferences, and capabilities of the agent, then the recognizer should combine this with evidence from the observed actions in reasoning about the entire plan.

Our framework is *Bayesian* in that we start from a causal theory of how the agent's mental state causes its plan and executing its plan causes activity, and reason from observed effects to underlying causes. Our recognizer has uncertain *a priori* knowledge about the agent's mental state, the world state, and the world's dynamics, which can be summarized (at least in principle) by a probability distribution. It then makes partial observations about the world, and uses this evidence to induce properties of the agent and its plan.

The remainder of this section describes our framework in more detail. We demonstrate the utility of the framework by showing how extensions to the underlying conception of plans and planning generate corresponding extensions to plan recognition. Examples from our explorations of plan recognition in a highway traffic domain illustrate our application of the framework to a concrete problem.

### 2.2 PLANNING MODEL

We begin with a model of the planning agent operating in the world. As it begins planning, the agent has a certain mental state, consisting of its preferences (e.g., goals), beliefs (e.g., about the state of its environment), and capabilities (e.g., available actions). We assume the actual planning process to be some rational procedure for generating the plan that will best satisfy the agent's preferences based on its beliefs, subject to its capabilities. This plan then determines (perhaps with some uncertainty) the actions taken by the agent in the world.

Most plan-recognition work concentrates only on this last step, the relationship between a plan and the actions taken in the world. Typical approaches start from a representation of the possible plans, and prune the set of possibilities based on the actions observed. For example, Kautz (1986) connects plans and actions through event hierarchies, which place the plan at the top of a taxonomy of subplans and actions. Vilain (1990) presents a context-free grammar representation of these event hierarchies as an alternative model. Lin and Goebel (1991) restrict the constraint language, permitting use of a faster, specialized message-passing recognition algorithm.

Given the reduced set of possible plans that could explain the observations, the plan recognizer must apply some preference criterion for choosing among them. For instance, Kautz's approach prefers explanations that involve fewer plans. The algorithm of Lin and Goebel prefers plan scenarios that are more general. However, given two explanations containing the same number of plans, at the same levels of generality, neither algorithm has a basis for a choice either way. To borrow an example from Charniak and Goldman, suppose we hear that Jack packed a bag and went to the airport. Depending on the exact event hierarchy, neither algorithm may be able to decide whether Jack is in the process of taking a trip or conducting a terrorist bombing.

The average reader would probably not consider the latter possibility, since people are much more likely to take a trip than bomb an airplane. Charniak and Goldman account for this behavior in their recognition procedure by including prior probabilities on plans. This allows them to distinguish among equally possible, but unequally plausible explanations for observed activity. The recognition model of Carberry (1990), based on the Dempster-Shafer theory of evidential reasoning instead of Bayesian techniques, takes a similar approach by using threshold plausibility and difference levels of belief to distinguish among competing hypotheses. Similar distinctions could be supported in linguistic approaches as well, perhaps based on probabilistic grammars (Wetherell, 1980).

### 2.3 MENTAL STATE

In a particular case, we typically have information available to us that would augment these prior probabilities. For instance, we may know that Jack belongs to a terrorist organization, which would make the bombing explanation of his actions more plausible. To account for this sort of knowledge, the plan-recognition framework should accommodate all possible information about the agent's plan selection process, beginning with its mental state. We can break down an agent's mental state into three distinct com-

ponents:

**Beliefs.** The agent's knowledge of the state of the world and its dynamics. Beliefs may be incomplete, uncertain, or incorrect.

**Preferences.** The agent's desires about the world. These may be simple goals, or arbitrarily graded degrees of utility.

**Capabilities.** The agent's self-model of its available actions. Strictly speaking, this should be *knowledge of* capabilities, but we stick to the more concise term.

We may have knowledge about any of these components of mental state. Looking back at Jack's situation, if we know that he belongs to a terrorist organization, then we might infer that his training included a lot of information about bombs, airport security, and other matters that are not widely known. Similarly, we may conclude that his goals are vastly different from those of a typical person going to the airport. For example, we may expect that Jack's goals include gaining worldwide attention for his group. Finally, his terrorist background may be such that he has a repertoire of available actions, such as conceal-bomb, beyond that of the vanilla air traveler.

Plan selection also relies on the agent's beliefs about the current world state. For instance, if Jack knows that there is an important diplomat on an outgoing flight, then he probably believes that bombing that plane will generate even more attention for his organization. Notice that the world state affects plan selection only through the agent's beliefs. If Jack did not know about any diplomats, then the fact that they are present is irrelevant to his planning. By the same token, if Jack believes that a diplomat is on the plane even if none are present, it is his erroneous belief that we must consider.

## 2.4   PLAN EXECUTION

Once we have accounted for the agent's plan-generation process, we need to consider the effects of the plan's execution. In many plan-recognition domains, the external observer finds the agent's actions inaccessible. In such cases, the recognizer observes actions only indirectly, via their effects on the world (which themselves are typically only partially observable). These restricted observations then form the basis of inference.

Thus, observations of the state of the world provide two types of evidence about the plan. First, as mentioned in Section 2.3, the world influences the agent's initial mental state, which provides the *context* for plan generation. Second, changes in the world state reflect the effects of the agent's actions, which *result* from executing its plan.

## 3   THE PLAN-RECOGNITION NETWORK

To perform plan recognition tasks, we generate a Bayesian network representing the causal planning model and use it to support evidential reasoning from observations to plan hypotheses. The structure of the Bayesian network is based on the framework depicted in Figure 1. That diagram can itself be viewed as a Bayesian network, albeit with rather broad random variables. To make this operational, we replace each component of the model with a subnetwork that captures intermediate structure for the particular problem. The limited connections among the subnetworks reflect the dependency structure of our generic planning model.

To illustrate this plan-recognition framework, consider the example problem of a driver on the highway, trying to predict the actions of the other drivers. Since these actions are normally limited to a small set of maneuvers (e.g. lane changes, passing, exiting), recognition of a driver's maneuvering plan would greatly assist in the prediction of future actions. To this end, we have worked on a probabilistic model of the maneuvers of a single car. We can then use this model to identify the current maneuver of an observed car and/or predict future actions, given only partial information. The subnetwork descriptions below first present the general construction techniques and then provide a specific instantiation for this specific traffic domain.

### 3.1   CONTEXT

The network, like the causal model, begins with the initial world state. We must include all possibly observable events that are relevant to *formation*—the process by which the agent's mental state is affected by the world. By including these events, the recognition procedure can take advantage of partial information about the agent's mental state. Note that even though the initial world state model may itself include inaccessible variables, the context subnetwork includes only those which are observable. However, we may wish to simplify the network by providing more compact intermediate results derived from inaccessible variables.

One of the motivations for maintaining a separate initial-state subnetwork is to distinguish between our contextual observations and those of the agent. Therefore, we may have an unobservable node representing an aspect of the world state accessible to the agent, and an observable node representing a related feature accessible to us. The dependency between these nodes is essentially a sensor model. If we are fortunate enough to have perfect sensors, then the context variables become redundant, since they will simply echo the values of the actual variables, and can be eliminated.

In this model, the initial world state is defined as causally prior to all agent behavior. Therefore, the corresponding random variables can have links only from other such variables, representing dependencies within the state. Any dependency links connecting a node from the initial state to any node outside this subnetwork must be directed to the outside node.

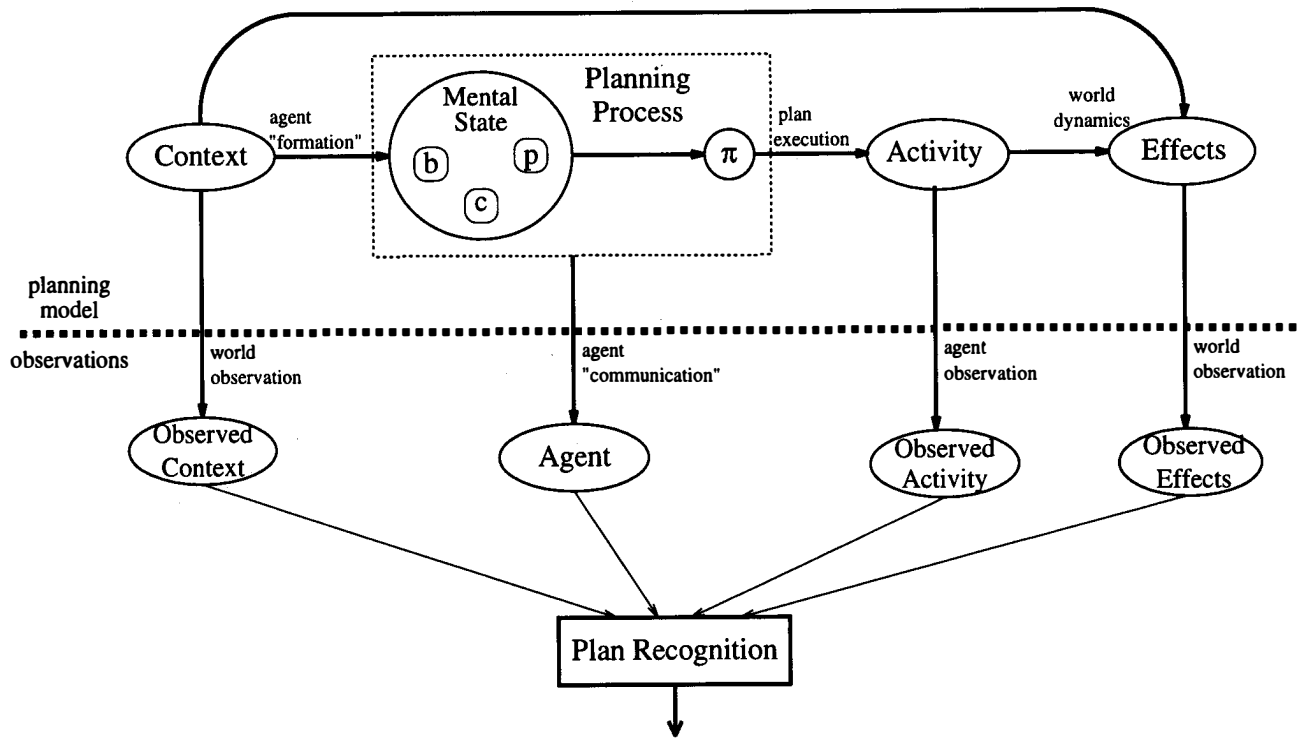

Figure 1: Plan Recognition Framework

This treatment of context differs from the work of Huber et al. (1994), where the initial situation depends on the agent's mental state and not the other way around as it is here. This was possible given the planning model employed in that work, that of the Procedural Reasoning System (PRS) (Ingrand et al., 1992). In the PRS model, plan selection is a function of current goal and situation. Because these context variables have no predecessors or substructure, the direction of links can be reversed without changing the rest of the dependency structure. However, the agent's mental state considered here may be more complex, especially in terms of its preference structure. Even if the agent has only simple goals, there are potential interactions among the goals that could affect the planning process. Hopefully, by following the causal structure in creating the network and placing the context prior to the plan, we can represent these interactions without greatly complicating the dependency structure.

In the traffic domain, the driver must consider several aspects of the initial world state in rationally choosing a plan. First of all, the current position and speed of the car are important factors, and we assume that both are observable, to the driver as well as to us. We also assume perfect sensors, but an extension to incorporate sensor noise is straightforward, as described above. The random variables x position and y position of Figure 2 represent the car's lane position and distance from the highway's start, respectively. The driver can be in one of three lanes or may be off the highway, either preparing to enter or having just exited. The random variable y speed, denoting the car's speed, initially depends on the current node, since the farther left the lane, the faster the car is usually traveling.

We can also observe the presence of other cars around the driver of interest, who must consider them in choosing a maneuver. For instance, if there is a car blocking the driver's front, then a passing maneuver is more likely. We can observe any cars to the driver's immediate front, back, left, and right, as well as in the four diagonal directions. In the Bayesian network, the Boolean random variable left clr? represents the presence of any car to the immediate left of the driver. There are similar variables for the right, front, and back, as well as the four diagonal directions. The variables indexed t0 in the first column of nodes in Figure 2 constitute the context subnetwork.

### 3.2 MENTAL STATE

The subnetwork representing the agent's beliefs about the world state must include random variables for all aspects of the context that the agent can observe and that factor into its decision-making. There may be some agent beliefs that are independent of any real-world variable. Unless we can observe these (perhaps through communication with the agent), there is no advantage in using additional random variables. Instead, we can fold the uncertainty in these beliefs into the plan subnetwork. However, agent beliefs will typically depend on the some aspect of the actual state of the world, although we can model the agent as being arbitrarily uncertain or deluded. As mentioned in Section 3.1, this dependency represents the imperfection and/or incompleteness of sensors. If the agent's sensors were perfect, then we could eliminate the nodes for the agent's belief variables, as they would take on the same values as the context variables.

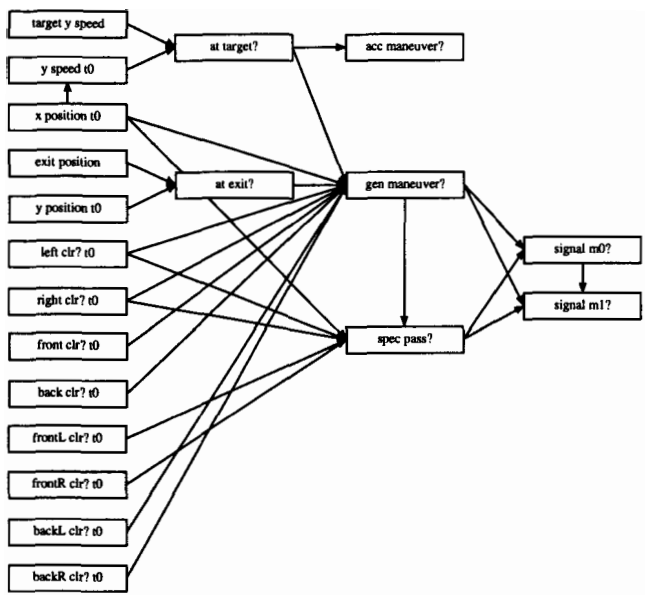

Figure 2: Planning process subnetwork

The agent's knowledge of its capabilities is usually independent of the world state, as are its preferences in most cases. Simple goals can be represented as separate Boolean variables, though it may be useful to combine a set of mutually exclusive variables into a single variable with several possible values. More complex preference structures will require more complex subnetwork structures. The agent's capabilities can be represented in a similar fashion.

The model of agent formation is greatly simplified in our traffic domain. Because of our assumption of perfect sensors, the driver's beliefs about the world correspond to the actual values in our simplified model. In addition, the agent's beliefs about its capabilities are not represented explicitly in our traffic network. Instead, the driver is assumed to know all of the possible plans (as described in Section 3.3.1). The planning process also assumes that the driver has complete knowledge of how the plans can best satisfy its preferences in the current context. Thus the plan selection mechanism implicitly represents the driver's beliefs about its capabilities.

We model the driver preferences with two goals. First, a driver has the explicit goal of getting from one exit to another, though the intended exits are unknown to an external observer. The random variable exit position in Figure 2 represents the driver's desired exit. All of the possible exit positions are farther along the highway than the values of y position. If this were not the case, then the current position would provide evidence that the desired exit is probably not one that has been passed. Therefore, there would be a dependency, but to simplify the network, we make the sets of $y$ and exit positions disjoint.

Second, there may be some constraint on the travel time between these exits, or the driver might have some target speed which is preferred for the duration of travel. However, we can usually translate the former into a desired speed because of the fixed positions of the exits. Therefore, our model uses only the random variable target y speed in Figure 2, with its values clustered around the speed limit. If the car has been on the highway for enough time, then its current speed should provide some clue as to the driver's target speed. We could model this with a link from y speed. On the other hand, if we have been observing the car and its maneuvers for some time, then these past observations should provide more conclusive evidence as to its target speed. Thus, we can make the target speed independent of current speed and encode our past observations in the prior probabilities.

This network also contains the intermediate belief random variables, at exit? and at target?, in the second column of nodes in Figure 2. These reflect the driver's belief about the proximity of the desired exit and the desirability of the current speed, respectively. The at exit? variable depends only on the current position and the preferred exit, and is true only when the former is immediately before the latter. The at target? variable depends only on the current and preferred speeds, and its value indicates whether the current speed is too slow, too fast, or just right, with respect to the driver's desired cruising speed.

### 3.3 PLANNING PROCESS

#### 3.3.1 Plan Variables

The plan subnetwork is comprised of random variables collectively representing the current plan. For instance, in Kautz's event hierarchies, there is a taxonomy of plans and actions. The children of a certain plan correspond to possible subplans or actions, while other links indicate necessary components. If our planning model is based on such event hierarchies, we may designate one Boolean variable corresponding to each element in the taxonomy, indicating the presence of the corresponding plan. Or we may combine certain mutually exclusive subplans into a single ran-dom variable, which takes on a different value depending on the actual subplan present.

Such hierarchies are based on the subsumption relation, requiring a dependency link from the more general node to the more specific. The conditional probability table can represent the distribution of the specific values, given the general. In particular, because of the subsumption relation, we can set the conditional probability of a child node given that its parent node is false to zero.

In the traffic domain, we can classify driving maneuvers according to the lane changes involved. The simplest plan is to simply continue driving in the same lane. At the next level of complexity, a driver can shift one lane to the left or right. We consider entering and exiting the highway as specific instances of these one-lane shifts. The driver could also shift two lanes to the left or right, where this could again involve entering or exiting the highway. As a final option, the driver may choose a passing maneuver, which we view as two successive lane shifts of opposite direction. In Figure 2, the variable **gen maneuver** represents the general driving maneuver and takes on a value corresponding to the chosen plan.

We can also classify driving plans according to the acceleration. Depending on the current and desired speed, a driver may decide to speed up, slow down, or maintain current speed, indicated by the value of the variable **acc maneuver** of Figure 2. The acceleration maneuver depends on the lane maneuver if we do not consider the plan selection mechanism. For instance, a deceleration is more likely as a part of a right lane change plan than as a part of a plan to pass. However, the two variables are independent given the initial context, as indicated in the network.

The variable **spec pass** in Figure 2 indicates the direction of the pass, if one is taking place. Since passing in a specific direction is a subplan of the general passing maneuver which **gen maneuver** can represent, this is an example of the subsumption relation found in event hierarchies. If the driver decides to pass, there are the options of passing on the left and passing on the right. And even if the driver chooses to pass, there may be cars blocking both lanes, forcing the driver to wait for another opportunity to pass. This variable clearly depends on **gen maneuver**, since the more general passing maneuver is its parent and the conditional probability table represents a subsumption relation as described above. In other words, if a passing maneuver is *not* chosen, then **spec pass** will be neither pass on left nor pass on right.

### 3.3.2 Plan Selection

Links from the agent's mental state into the plan subnetwork represent the agent's planning process. For hierarchical planning, we start with the most general plan nodes and proceeding to the most specific, determine which aspects of the mental state influence the agent's choice. For instance, suppose the agent's decision-making procedure consists of a set of condition-action rules. Then, any plan choices in the action portion of a rule depend on all of the context variables that appear in the conditions of the rule. By connecting only parts of the mental state relevant to particular choices, we keep the dependency structure as simple as possible.

We must then specify the conditional probabilities of the plan variables given the relevant aspects of the agent's mental state. If the agent is a deterministic planner, then the conditional probability given a particular mental state instantiation will be 1 for a single instantiation of the plan subnetwork and 0 for all others. For nondeterministic planners, we must determine the conditional probabilities from whatever agent model we have.

If in fact we have no opportunity to observe anything about the initial world state or the agent's mental state, then we may collapse the initial state and mental state subnetworks into prior probabilities for the top-level plan variables. The plan recognition networks (PRNs) of Charniak and Goldman (1993) use such priors to model the agent's plan selection process. These prior probabilities represent the same distribution as the explicit planning process subnetwork, but since the initial nodes are unobservable, we can merge the nodes into the plan subnetwork without losing information.

We can now model a driver's plan selection with some reliability. In our Bayesian network, the conditional probability table must specify the likelihood of certain maneuvers under every possible combination of world situation and driver mental state. Under most situations, there will be one maneuver that is clearly preferable, though there is still uncertainty. For example, suppose that the driver is currently traveling below the desired speed and that there is another car directly in front while the lane to the left is clear. Then it is likely that driver will pass the car on the left. The complete plan selection subnetwork is shown in Figure 2. This model of the driver's decision process is based in part on the driving model underlying the BATmobile (Bayesian Automated Taxi) project, described by Forbes, et al. (1995).

The acceleration maneuver depends only on the preferability of the current speed. Thus the sole link to **acc maneuver** is from **at target?**. If the driver is at the target speed, then the current speed will be maintained. If the current speed is too low, then the driver will choose an acceleration maneuver. Likewise, if the current speed is too fast, then a deceleration maneuver will be chosen.

The lane change maneuver also depends on the preferability of the current speed. For instance, a car traveling at its target speed is unlikely to change lanes. However, there are other factors in the initial world state to consider. Obviously, the current lane is important, since a car in the leftmost lane cannot change lanes to the left. In addition, the driver will consider any cars to the front or back. If there is a car blocking the front and the driver's current speed is too low, then a simple acceleration could cause a collision. The driver may instead choose to change lanes to the left. But a decision to change lanes must also consider the presence of cars to the driver's left or right, or any cars coming up from the back left or right. The links to the gen maneuver node represent these dependencies.

If the driver decides to pass, a direction must be chosen. Passing on the left is preferable to passing on the right, but the current situation may not allow it. For instance, any cars to the driver's left or to the front left could block the passing attempt. The same is true on the right side. If enough passing avenues are blocked, then the driver may decide to delay the passing attempt or to perform the initial lane change and wait to complete the pass.

### 3.3.3 Agent Communication

Modeling agent communication depends greatly on the specific protocol adopted, and the relationship between the observer and the observed. If a trusted agent directly announces particular aspects of its planning process, then we could simply instantiate the corresponding variables. Other types of communication would require nodes to represent beliefs we attribute to the agent, based on its communication actions. Note that we are not modeling here the planned character of communication acts; to do so we would treat them as we do actions in general.

The only communication allowed in our traffic model is through the driver's turn indicator, which provides a simple mechanism for a driver to announce the intended lane change. The variables signal m$x$? in the fourth column of Figure 2 represent the state of the driver's turn signal during stage $x$ of the maneuver. Clearly, both the general maneuver and the specific direction of any passing attempt influence any signal. For instance, when performing a left lane change, signal m0? is likely to take the value *Left* and signal m1? the value *Off*. Of course, many drivers fail to signal their maneuvers, and sometimes they signal erroneously. These possibilities are considered when determining the conditional probability tables. However, drivers are usually consistent in their signaling habits. For instance, when performing a pass on the left, someone who fails to signal the initial left lane change is unlikely to signal the subsequent right change. The link between the two signal variables represents this consistency.

### 3.4 PLAN EXECUTION

The agent's plan execution process is reflected in the model by dependencies from its plan subnetwork to another subnetwork describing its activity. This is analogous to links in event hierarchies connecting plans to their component ob-servable actions. In PRS (Ingrand *et al.*, 1992; Lee *et al.*, 1994), Knowledge Areas (KAs) specify a sequence of actions associated with a plan, corresponding to links from the plan node to corresponding action nodes. Either of these can be cast in Bayesian networks, representing the likelihood of the component's appearance given the plan in the conditional probability table for that node.

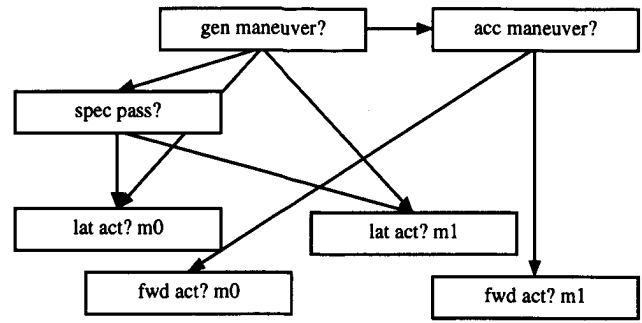

Figure 3: Plan execution subnetwork

All of these methods for modeling the dependency of the agent's activity on its plan are acceptable. We require only that the agent's activity be conditionally independent of the initial world state and the agent's mental state given the plan. That is, we assume that the plan is a sufficient specification of activity.

The activity subnetwork in the traffic model includes the individual transitions in lane and speed, which are completely unobservable. At each step, the driver can change one lane to the left or right, or remain in the same lane. The driver can also increase, decrease, or maintain speed. All of the plans we consider produce a two-step action sequence. For instance, a plan to shift one lane to the left produces a left lane change followed by a "remain in lane" act. The lat act m$x$ variables in Figure 3 represent the lane changes at step $x$, while fwd act m$x$ represents the acceleration at step $x$.

Our definition of the lane maneuvers completely determines the lane changes of the action sequences. The individual shifts depend on the general lane maneuver, as well as on the specific passing plan, but not on the acceleration maneuver. Likewise, the individual accelerations are independent of the general lane changes and the specific passing maneuvers if given the overall acceleration plan.

### 3.5 WORLD DYNAMICS

The relationship between the observed and actual actions of the agent is similar to that of the observed and actual world states. If we have perfect sensors, we do not need a separate observed activity subnetwork; otherwise, we have to model sensor noise in the links from the actual nodes.

In some cases, the agent's activity is completely inaccessi-ble, though we might still be able to observe effects of this activity. These effects are dictated by the dynamics of our world, which specify how the agent's actions alter the situation. Therefore, we must model how subsequent world states depend on the initial world state and the agent's activity. It is possible that a world state depends on the entire world history, but if the the plan is sufficiently structured (e.g., sequential actions) then we may be able to simplify this dependency. If we express the effects model in accord with standard AI approaches, we can restrict the effects to depend only on background and direct effects and, given these, to be conditionally independent of the plan itself, as well as further removed activity and indirect effects (Wellman, 1990).

We can make effects conditionally independent of future actions and effects simply by ensuring that links never point backward in time, but this could make actions dependent on past world states. So far, we have had links move from plans to activity and from activity to effects, so adding links in the opposite direction would go against the flow in Figure 1. If, as described above, the plan is sufficient for determining activity, the current action is conditionally independent of the previous world states given the current plan, as well as the actions performed so far.

Depending on our domain, we may able to make a Markov assumption with respect to activity and the effects. In such cases, the current action would be conditionally independent of actions more than one time step back in the action sequence, given the current plan and the action immediately previous. If the effects have a similar property, they should not depend on any world states or actions more than one step previous. Although this would greatly localize the dependencies, this may not always be possible, depending on the types of observations available and the set of state variables in the model.

Since there is no directly observable activity in the traffic model, most of our inference will come from observed effects. We must now model the dynamics of the traffic world, beginning with the changes in the position and speed of the car. We can view the actions of the driver to be transitions between world states. To simplify the model, we ignore observations taking place while the driver is performing an action. Thus, evidence is available only at the completion of a component action, and there are three stages of observable variables, including the context, as can be seen in Figure 4.

Finally, we must define the dependencies of these effects. Most of the observable variables depend on the driver's previous action, as well as their own previous values. For instance, the driver's lane is completely determined if we know what lane change just took place, as well as the lane value just before the change. Likewise, the driver's speed depends on the previous speed and whatever acceleration action took place, although this is clearly not a deterministic relationship.

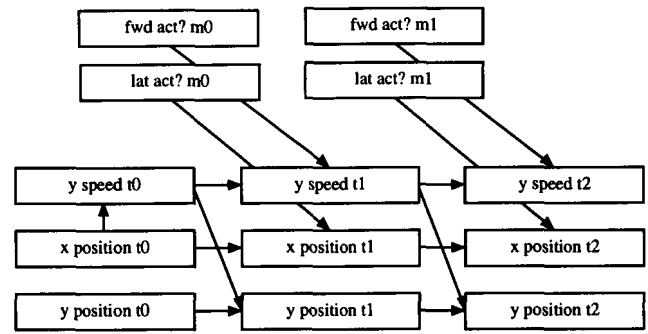

Figure 4: Observation subnetwork

The presence of other cars is a bit more complex, due to the driver's movements. For instance, after a left change, a car that was to the front and left is now probably directly in front. But if the driver stays in the same lane, then we must check whether there was a car blocking the front in the previous world state. Therefore, each clearance variable depends on the previous action, as well as all relevant clearance variables from the previous state. To simplify the network, we ignore the presence of other cars in the evidence. We do consider them when modeling plan selection, but since the driver's actions do not directly affect the other drivers' positions, we ignore these effects. As with the context, we assume perfect sensors, so there is no distinction between the actual and observed effects.

### 3.6 PLAN RECOGNITION

Once we have created the belief network, we can perform recognition tasks by fixing any observed variables and querying the network about the relevant variables. We receive evidence only about the variables in the bottom half of Figure 1, though, as described before, these may correspond exactly to actual variables in the planning model.

Once we fix the values of the known variables in the network, we can propagate the information throughout the network and observe the posterior probabilities at the nodes of interest. For instance, we may be interested in determining the plan chosen by the agent, in which case we would examine the nodes in the plan subnetwork. Alternatively, we can predict future agent activity or effects by examining the probabilities of those variables.

Once we have constructed the entire traffic maneuver network, shown in Figure 5, we can handle plan recognition in a wide range of useful driving situations. For instance, suppose we are trying to predict the behavior of the car behind us as we are driving in the middle lane of a three-lane highway. We observe the car move into the rightmost lane, and we want to determine if it is passing us, or preparing to exit,

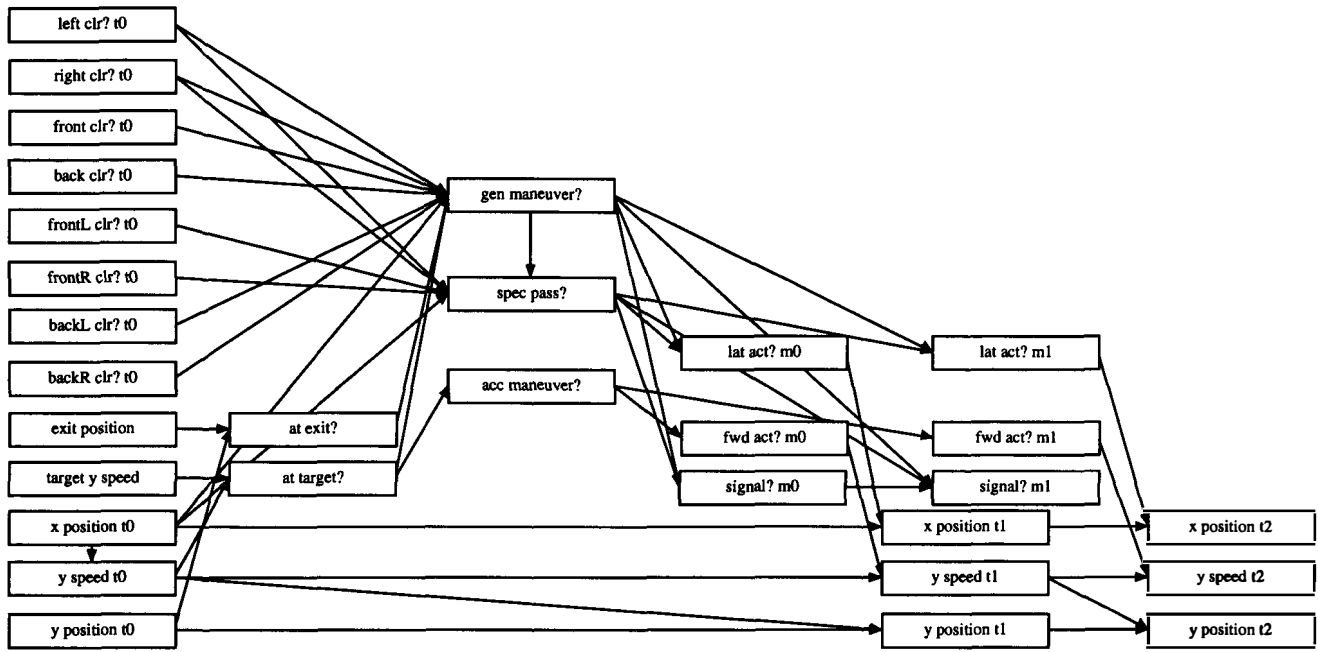

Figure 5: Complete Bayesian network for traffic monitoring

or perhaps simply moving into the slower-moving lane.

Thus, in the context, we have observed front clr? t0 to be false and x position t0 to be the middle lane. The only observed effect is that x position t1 is the right lane. If we want to infer the driver's plan, we can examine the gen maneuver? node to see that the posterior probability of a one-lane right shift is 0.64, while that of a pass is 0.35. The former is more plausible since we assume that drivers prefer to pass on the left-hand side, so passing on the right has a relatively low prior probability. The only remaining maneuver with nonzero probability is an exit. All of the other plans have zero probability, since the observed change in lanes violates their definitions.

If we are not interested in the driver's plan, but only in the future lane position, then we can examine the x position t2 node. The posterior probability that the car will still be in the right lane is 0.65, while the probability that it will move to the middle lane is 0.34. The difference between these beliefs and that of the maneuvers arises from the nature of the passing maneuver. Even if the car decides to pass, it may not be able to do so immediately do to surrounding cars. In such a case, it will remain in its current lane until it can complete the maneuver. Thus, there is a slight probability that the car will stay in the right lane even if the driver has decided to pass.

Given no other contextual observations, it is reasonable to predict that the car will remain in the right lane. However, if we also observed that there was another car to our left, thus blocking the car behind us from passing on the left, we can instantiate the frontL clr? t0 variable to be false. Repeating our observation of the nodes of interest, we find that the posterior probability that the car is passing has increased to 0.53, while that for the car simply shifting one lane to the right has dropped to 0.46. The probabilities for x position t2 have changed as well, to 0.51 and 0.48 respectively. If we made our final decisions based simply on maximum probabilities, we would predict that the car was passing us. Notice that, without knowing about the car to our left, our prediction would be that the car was not passing, but the observation of that aspect of the context changes our belief.

Thus, we are able to perform valuable inference with only a limited subset of the possible observations. If we were to also observe that there were no other cars nearby, other than those already considered, then we could instantiate the remaining clearance context variables to be true. Doing so increases the posterior probability that the maneuver is a pass to 0.61, while decreasing that for a one-lane right shift to 0.39.

## 4 CONCLUSION

The traffic application presented above illustrates several aspects of our plan-recognition framework, highlighting the importance of accounting for context. Our assumption of rationality on the part of the agent allowed us to model the relationship between an agent's plan and its mental state. By modeling a driver's decision process, observations of the initial state provided strong evidence about the resulting plan. We were also able to model plan execution in a manner similar to other approaches to recognition. The resulting network was able to perform useful inference, even when given only partial observations.

Although the traffic example is a very specific domain, we believe that the general structure of Figure 1 is applicable to a broad class of plan-recognition tasks. Even with our restrictive assumptions, the network captures an extensive model of planning behavior. The driver observes the world, generates a plan, performs a sequence of actions, and these actions produce changes in the world state. To summarize, we have augmented the common plan execution model employed in recognition to include plan formulation, and the result is encouraging.

However, the generality and scalability of our framework remains to be seen. The driver we considered had two goals, an intended exit and a target driving speed. Other drivers, and agents in other domains, will most likely have more complex preferences and a more complex decision process. The decision process may involve a more elaborate planning theory, which may be difficult to capture in our model. In addition, the major issue of communication is as yet unexplored within our model. In future work, we intend to push on these issues, by increasing the scale and complexity of the underlying process we are modeling.

## Acknowledgments

This work was supported in part by Grant F49620-94-1-0027 from the Air Force Office of Scientific Research, and by the Transportation Research Board IVHS-IDEA Program. Marcus J. Huber provided many helpful suggestions in discussions of this work. The authors also thank the anonymous reviewers for their comments on the exposition.